\documentclass[10pt,twocolumn,letterpaper]{article}

\usepackage{cvpr}
\usepackage{times}
\usepackage{epsfig}
\usepackage{graphicx}
\usepackage{amsmath}
\usepackage{amssymb}

\usepackage{xcolor}

\usepackage{booktabs}

\usepackage{multirow}
\usepackage{array}
\usepackage{algorithm}

\cvprfinalcopy 

\def\cvprPaperID{666} 

\ifcvprfinal\fi
\begin{document}

\title{MVF-Net: Multi-View 3D Face Morphable Model Regression}

\author{
Fanzi Wu$^{2}$\footnotemark[2]\hspace{4pt}\footnotemark[1]\quad 
Linchao Bao$^{1}$\footnotemark[1] \quad 
Yajing Chen$^{3}$ \quad 
Yonggen Ling$^{1}$ \\
Yibing Song$^{1}$ \quad
Songnan Li$^{2}$ \quad 
King Ngi Ngan$^{2,4}$ \quad 
Wei Liu$^{1}$ \quad \vspace{5pt}\\
$^{1}$Tencent AI Lab \quad
$^{2}$The Chinese University of Hong Kong \\
$^{3}$Shanghai Jiao Tong University\quad
$^{4}$University of Electronic Science and Technology of China\quad\\}

\maketitle
\thispagestyle{empty} 

\footnotetext[1]{Equal contributions. Emails: \{wufanzi412, linchaobao\}@gmail.com}
\footnotetext[2]{This work was done when Fanzi Wu was an intern at Tencent AI Lab.}
\begin{abstract}
We address the problem of recovering the 3D geometry of a human face from a set of facial images in multiple views.
While recent studies have shown impressive progress in 3D Morphable Model (3DMM) based facial reconstruction, the settings are mostly restricted to a single view. 
There is an inherent drawback in the single-view setting: the lack of reliable 3D constraints can cause unresolvable ambiguities. 
We in this paper explore 3DMM-based shape recovery in a different setting, where a set of multi-view facial images are given as input. 
A novel approach is proposed to regress 3DMM parameters from multi-view inputs with an end-to-end trainable Convolutional Neural Network (CNN).
Multi-view geometric constraints are incorporated into the network by establishing dense correspondences between different views leveraging a novel self-supervised view alignment loss. 
The main ingredient of the view alignment loss is a differentiable dense optical flow estimator that can backpropagate the alignment errors between an input view and a synthetic rendering from another input view, which is projected to the target view through the 3D shape to be inferred. 
Through minimizing the view alignment loss, better 3D shapes can be recovered such that the synthetic projections from one view to another can better align with the observed image. 
Extensive experiments demonstrate the superiority of the proposed method over other 3DMM methods.

\end{abstract}

\section{Introduction}

Reconstructing 3D facial shapes from 2D images is essential for many virtual reality (VR) and augmented reality (AR) applications. 
In order to obtain fully-rigged 3D meshes that are necessary for subsequent steps like facial animations and editing, 3D Morphable Model (3DMM) \cite{blanz1999morphable} is often adopted in the reconstruction to provide a
parametric representation of 3D face models. 
While conventional approaches recover the 3DMM parameters of given facial images through analysis-by-synthesis optimization \cite{blanz2003face,romdhani2005estimating}, recent work has demonstrated the effectiveness of regressing 3DMM parameters using convolutional neural networks (CNN) \cite{zhu2016face,tran2017regressing,tewari2017mofa,kim2018inversefacenet,genova2018unsupervised, shen2018person, shen2018deep}. In spite of the remarkable progress in this topic, recovering 3DMM parameters from a single view suffers from an inherent drawback: the lack of reliable 3D constraints can cause unresolvable ambiguities, \eg, the height of nose and cheekbones of a face is difficult to tell given only a frontal view. 

\begin{figure}
\includegraphics[width=\linewidth]{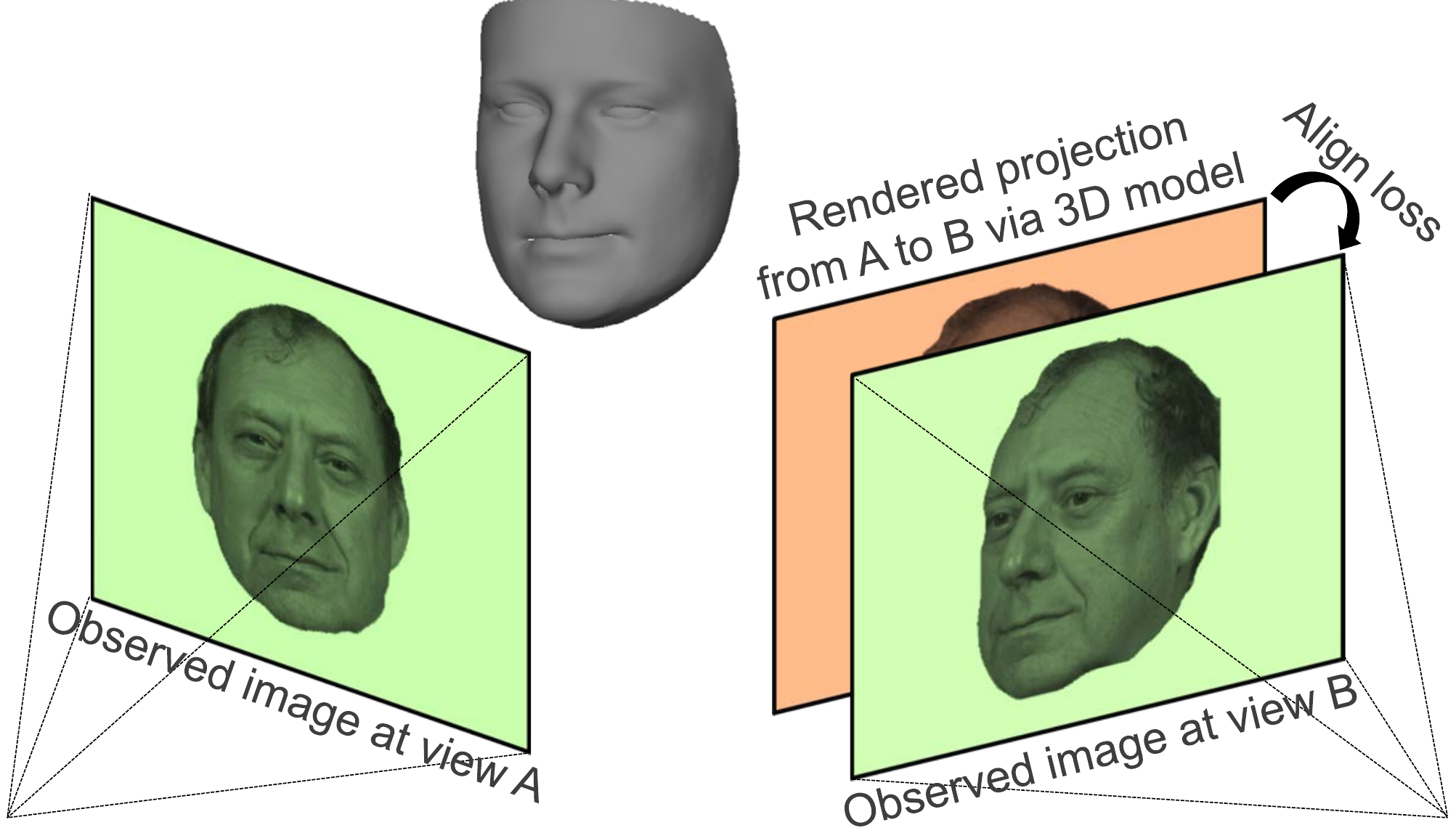}
\caption{An illustration of the view alignment loss. The rendered projection from view A to B via the optimal underlying 3D model should align best with the image observed at view B.}
\label{fig:multiview3dface}
\vspace{-10pt}
\end{figure}

A better way to reconstruct more faithful 3D shapes from 2D images is to exploit multi-view geometric constraints using a set of facial images in different views. In this case, structure-from-motion (SfM) and multi-view stereo (MVS) algorithms \cite{furukawa2015multi} can be employed to reconstruct an initial 3D model and then a 3DMM fitting can be performed using the 3D geometric constraints from the initial model \cite{blanz1999morphable}. However, the separated two steps are error-prone: the SfM/MVS step cannot utilize the strong human facial prior from 3DMM and hence its results are usually rather noisy, which further leads to erroneous 3DMM fitting. An alternative approach is to directly fit 3DMM parameters from multi-view images through analysis-by-synthesis optimization \cite{romdhani2005estimating}, but it requires a complicated, nonlinear optimization that can be difficult to solve in practice. 

In this paper we propose a novel approach, which adopts an end-to-end trainable CNN to regress 3DMM parameters in the multi-view setting. 
Inspired by the photometric bundle adjustment method \cite{delaunoy2014photometric} for camera pose and 3D shape estimation in multi-view 3D reconstruction, our method is also based on the assumption that the underlying optimal 3D model should best explain the observed images in different views. That is, the photometric reprojection error between each observed image and a rendered image induced by the underlying 3D model for this view should be minimized  (as illustrated in Fig. \ref{fig:multiview3dface}). 
To incorporate this constraint into our CNN, we sample textures from an input view using the predicted 3D model and camera pose, and then render the textured 3D model to another view to compute the loss between the rendered image and the observed image in the target view. In addition to the direct photometric loss between the two images, we propose a novel view alignment loss utilizing a differentiable dense optical flow estimator to backpropagate alignment errors, to avoid trapping into local minima during training. 
All the above procedures are differentiable and the whole network is end-to-end trainable.
To the best of our knowledge, this is the first work that proposes an end-to-end trainable network to exploit both 3DMM and multi-view geometric constraints. We conduct extensive experiments to show the effectiveness of the proposed method.

\section{Related Work}

In this section, we briefly summarize the most related work to our approach.
Please refer to the recent survey \cite{zollhofer2018state} for more detailed review. 

\subsection{Morphable 3D Face Model (3DMM)} 
Blanz and Vetter \cite{blanz1999morphable} introduced the 3D morphable model to represent textured 3D faces using linear combinations of a set of shape and texture bases, which is derived from collections of real 3D face scans. The model is later extended to include facial expressions by FaceWarehouse \cite{cao2014facewarehouse}. In this paper, we focus on recovering the underlying 3D shapes of human faces, hence we are only interested in regressing 3DMM parameters for shapes and expressions. We argue that more realistic textures for 3D meshes can be obtained with more advanced texture synthesis techniques \cite{saito2017photorealistic} instead of the 3DMM texture representations.

\subsection{Single-view 3DMM-based Reconstruction} 

Conventional methods for single-view 3DMM fitting are mostly based on analysis-by-synthesis optimization \cite{blanz2003face,romdhani2005estimating,garrido2013reconstructing,thies2016face2face, wu2016model, wu20173d}, by constraining the data similarities like pixel colors, facial landmarks, edges, \etc, between observed images and the synthetic images induced by 3DMM. The optimization is usually sensitive to initial conditions and parameters, and hence brittle in practice. This leads to the recent interests in regression-based approaches with deep neural networks. 

Zhu \emph{et al.} \cite{zhu2016face} proposed a cascaded CNN to regress and progressively refine 3DMM parameters, trained with supervision data generated by fitting 3DMM parameters using conventional approaches and then augmented by their proposed face profiling technique. Later, Tran \etal \cite{tran2017regressing} presented that more discriminative results could be obtained with deeper networks and 3DMM pooling over face identities. However, both methods require supervision obtained through optimization-based 3DMM fitting techniques. Dou \etal \cite{dou2017end} proposed to train the regression network using real 3D scans together with synthetic rendered face images with a 3D vertex distance loss. Richardson \etal \cite{richardson20163d} showed that a 3DMM regression network can be trained using only synthetic rendered face images and later Kim \etal \cite{kim2018inversefacenet} proposed a bootstrapping algorithm to adapt the synthetic training data distribution to match real data. Recently, Tewari \etal \cite{tewari2017mofa} and Genova \etal \cite{genova2018unsupervised} demonstrated impressive results by training 3DMM regression networks using only unlabeled images with a self-supervised photometric loss and a face recognition loss, respectively. 

To model detailed facial geometries beyond the representation power of 3DMM, some recent studies proposed to supplement additional geometric representations such as displacement maps \cite{richardson2017learning,tran2018extreme} or parametric correctives \cite{Tewari_2018_CVPR} besides 3DMM representations. 
Some other work used volumetric representations \cite{jackson2017large} or non-regular meshes \cite{sela2017unrestricted} instead of parametric representations. These types of representations are out of the scope of this paper.

\subsection{Multi-view 3DMM-based Reconstruction} 

In the multi-view setting, a straightforward solution \cite{ichim2015dynamic} for 3DMM-based reconstruction is to first perform traditional multi-view 3D reconstruction \cite{furukawa2015multi} and then fit a 3DMM using the reconstructed 3D model as constraints. However, the separated two steps are error-prone: the SfM/MVS step cannot utilize the strong human facial prior from 3DMM and hence its results are usually rather noisy, which further leads to erroneous 3DMM fitting. Dou \etal \cite{dou2018multi} recently proposed to address the problem using deep convolutional neural networks (CNNs) together with recurrent neural networks (RNNs). They used RNNs to fuse identity-related features from CNNs to produce more discriminative reconstructions, but multi-view geometric constraints are not exploited in their approach. Notice that there are some other 3DMM-based methods in multi-image settings \cite{piotraschke2016automated}, but in these work each input image is dealt individually, which is not the same as our multi-view setting. 

\section{Approach}

\begin{figure*}
\includegraphics[width=\linewidth]{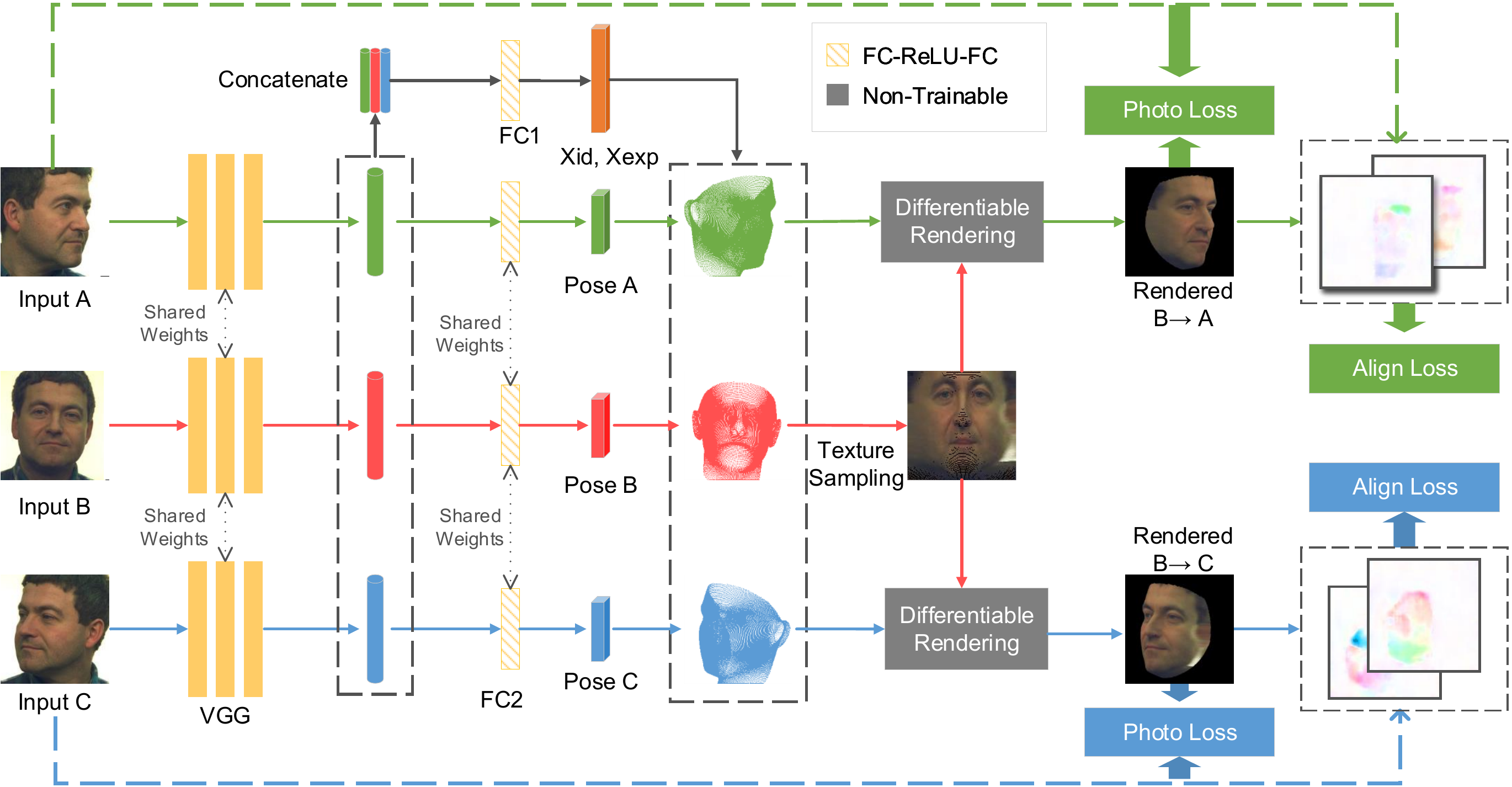}
\caption{An overview of the proposed model.}
\label{fig:overviewarch}
\vspace{-10pt}
\end{figure*}

\subsection{Overview}

We employ an end-to-end trainable CNN to regress 3DMM parameters from multiple facial images for the same person in different views. In order to establish multi-view geometric constraints like conventional multi-view 3D reconstruction approaches \cite{furukawa2015multi}, for now we assume the facial images are taken at the same time under the same lighting condition. Later we will illustrate that our approach is able to handle inputs with lighting variance. For simplicity, we adopt three-view setting to describe our approach. Note that it can be easily generalized to other number of input views. 

Fig. \ref{fig:overviewarch} illustrates the overview of our proposed model in the case of three input views. We learn features from each input image by a shared weight CNN, and then concatenate the features together to regress a set of 3DMM parameters for the person. Differently, we regress the pose parameters for each input view from its individual features (Sec. \ref{sec:paramregress}). With the pose parameters and 3DMM parameters, we are able to render a textured 3D face model from each input image by sampling textures from the image (Sec. \ref{sec:texturesample}). Note that in the three-view setting, there will be three textured 3D face models, with the same underlying 3D shape but with different textures. After obtaining the rendered 3D face models of different views, we then project each of them to a different view from the view where the textures are sampled (Sec. \ref{sec:renderproj}). For instance, we project the 3D model with textures sampled from image at view A to view B. Then we can compute losses between the projected image with the input image at the target view. We will present the details of the adopted losses in Sec. \ref{sec:losses}. 
Please be noted that the rendering layer is non-parametric yet differentiable, like that in previous self-supervised approaches \cite{tewari2017mofa,genova2018unsupervised}, and the gradients can thus be backpropagated to the trainable layers. 

\vspace{-1pt}
\subsection{Model}\label{sec:3dmmmodel}
\vspace{-3pt}

The 3DMM parameters to be regressed in this work include both identity and expression parameters like \cite{zhu2016face}. A 3D face model $\mathbf{s}$ can be represented as 
\begin{equation}\label{eq.3dmm}
  \mathbf{s} = \mathbf{\Bar{s}} + E_{\text{id}} \mathbf{x}_{\text{id}} + E_{\text{exp}} \mathbf{x}_{\text{exp}},
\end{equation}
where $\mathbf{\Bar{s}}$ is the vector format of the mean 3D face model, $E_{\text{id}}$ and $E_{\text{exp}}$ are the identity basis from BFM 2009 \cite{paysan20093d} and expression basis from FaceWarehouse \cite{cao2014facewarehouse} respectively, $\mathbf{x}_{\text{id}}$ and $\mathbf{x}_{\text{exp}}$ are the corresponding 199-dimension identity vector and 29-dimension expression vector to be regressed. 

To project 3D model onto 2D image plane, we employ the weak perspective projection model. Given a 3D point $\mathbf{v}$, its 2D projection can be computed with a set of camera pose parameters $\mathcal{P}$ as follows
\begin{equation}\label{eq.projectionmodel}
  \mathrm{Pr}(\mathbf{v}, \mathcal{P}) = 
  \begin{bmatrix}
    f & 0 & 0 \\
    0 & f & 0 
  \end{bmatrix} \cdot R \cdot \mathbf{v} + \mathbf{t},
\end{equation}
where $f$ is the scaling factor, $R$ is the rotation matrix, and $\mathbf{t}$ is the 2D translation $[t_x, t_y]^\text{T}$. Since the rotation matrix $R$ can be minimally parameterized as three Euler angles $\alpha$, $\beta$, $\gamma$, the pose to be regressed contains 6 parameters in total, which reads as $\mathcal{P}=\{f, \alpha, \beta, \gamma, t_x, t_y\}$. 

\subsection{Parametric Regression}\label{sec:paramregress}

We denote the three-view input images as $\mathbf{I}_A$, $\mathbf{I}_B$, and $\mathbf{I}_C$. We assume $\mathbf{I}_B$ is the image taken from the frontal view, $\mathbf{I}_A$ and $\mathbf{I}_C$ are taken from the left and right views respectively. Note that we do not need the images to be taken from precise known view angles. 
Each input image is sent through several convolutional layers (borrowed from VGG-Face \cite{Simonyan15vgg} in our implementation) and pooled to a 512-dimentional feature vector. Then a set of pose parameters $\mathcal{P}=\{f, \alpha, \beta, \gamma, t_x, t_y\}$ is regressed for each view via two fully-connected layers. 
The three 512-dimentional feature vectors are concatenated together to regress the 228-dimentional 3DMM parameters $\mathcal{X}=\{\mathbf{x}_{\text{id}}, \mathbf{x}_{\text{exp}}\}$ (199 for identity and 29 for expression) using another two fully-connected layers. Note that for each set of inputs, we regress one $\mathcal{X}$ and three pose parameters  $\mathcal{P}_{A}$, $\mathcal{P}_{B}$, and $\mathcal{P}_{C}$. The networks to extract features and regress pose parameters for the three views have shared weights. 

\subsection{Texture Sampling}\label{sec:texturesample}

With the predicted 3DMM parameters $\mathcal{X}$, as well as the known identity basis $E_\text{id}$ and expression basis  $E_\text{exp}$, we can compute the 3D face model using Eq. \eqref{eq.3dmm}. Three different texture maps can be obtained by sampling textures from each image individually using its own pose parameters predicted by the network. For each vertex $\mathbf{v}$ of the 3D model, we apply Eq. \eqref{eq.projectionmodel} to project the vertex to the image plane and fetch the texture color from each input image for the vertex using differentiable sampling scheme, as adopted in Spatial Transformer Networks \cite{jaderberg2015spatial}. For 3D point within a triangle on the mesh, we utilize barycentric interpolation to get its texture color from surrounding vertices. Note that since the texture sampling scheme does not handle occlusions, the textures sampled for occluded regions in each image are erroneous. We deal with this problem using visibility masks which will be detailed in Sec. \ref{sec:vismask}. Suppose now we have obtained three differently textured 3D models in this step.

\subsection{Rendered Projection and Visibility Masks}\label{sec:renderproj}\label{sec:vismask}

The textured 3D model can be projected to an arbitrary view to render a 2D image, via the differentiable rendering layer introduced in \cite{genova2018unsupervised}. For example, given a 3D model with textures sampled from image $\mathbf{I}_A$, we can render it to the view of $\mathbf{I}_B$ using the pose parameters $\mathcal{P}_{B}$, which we denote as $\mathbf{I}_{A \rightarrow B}$. Formally, for any 3D point $\mathbf{v}$ on the mesh surface (including points within triangles), the color of its projected pixel in the rendered image can be computed as
\begin{equation}\label{eq.renderproj}
  \mathbf{I}_{A \rightarrow B}[\mathrm{Pr}(\mathbf{v}, \mathcal{P}_{B})] = \mathbf{I}_A[\mathrm{Pr}(\mathbf{v}, \mathcal{P}_{A})],
\end{equation}
where we use $[\cdot]$ to denote the pixel selection in an image. 
In practice, the rendering is implemented through rasterization on the target image plane, that is, denoting an arbitrary pixel in the target image as $\mathbf{u}$, then Eq. \eqref{eq.renderproj} can be written as
\begin{equation}\label{eq.rasterize}
  \mathbf{I}_{A \rightarrow B}[\mathbf{u}] = \mathbf{I}_A[\mathrm{Pr}(\mathrm{Pr}^{-1}(\mathbf{u}, \mathcal{X},  \mathcal{P}_{B}), \mathcal{P}_{A})],
\end{equation}
where we use $\mathrm{Pr}^{-1}(\cdot)$ to denote the back projection from a 2D point to 3D space. Note that since the back projection is essentially a ray in 3D space, we need the 3D surface of the face model, which can be induced by 3DMM parameters $\mathcal{X}$, in order to locate the back projection ray to a 3D point. Thus the back projection operator $\mathrm{Pr}^{-1}(\cdot)$ in the above equation takes $\mathcal{X}$ as input in addition to camera pose $\mathcal{P}_{B}$. Ideally, with the optimal underlying 3D model and camera poses, the observed image $\mathbf{I}_B$ should be the same as the rendered image $\mathbf{I}_{A \rightarrow B}$ in non-occluded facial regions, 
\begin{equation}\label{eq.consistphoto}
  \mathbf{I}_{A \rightarrow B} (\mathcal{X}^*, \mathcal{P}^*_{B}, \mathcal{P}^*_{A})[\mathbf{u}] \equiv \mathbf{I}_B[\mathbf{u}], \text{~~for~} \mathbf{u} \in \mathcal{M},
\end{equation}
where $\mathcal{M}$ denotes the set of pixels in non-occluded facial regions. 
We will use this assumption to design our self-supervised losses in Sec. \ref{sec:selfsup}.

\begin{figure}[!t]
\centering
\begin{tabular}{c@{\hspace{0mm}}cccc}
   &\includegraphics[width=0.1\textwidth]{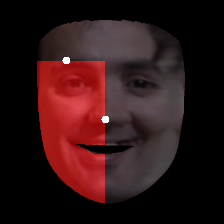}
   &\includegraphics[width=0.1\textwidth]{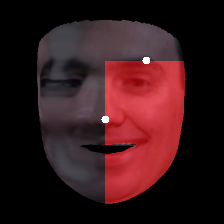}
   &\includegraphics[width=0.1\textwidth]{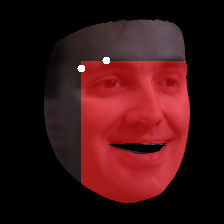}
   &\includegraphics[width=0.1\textwidth]{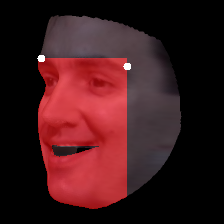}\\
   &\footnotesize{(a)} &\footnotesize{(b)} &\footnotesize{(c)} &\footnotesize{(d)}
\end{tabular}
   \caption{Visibility masks for rendered images:
   (a) $\mathbf{I}_{A \rightarrow B}$;
   (b) $\mathbf{I}_{C \rightarrow B}$;
   (c) $\mathbf{I}_{B \rightarrow A}$;
   (d) $\mathbf{I}_{B \rightarrow C}$. The dark regions are excluded using 3D landmarks on nose tip and eyebrows (the white points).}
\label{fig:masks}
\end{figure}

\begin{figure}[!t]
\centering
\begin{tabular}{c@{\hspace{0mm}}ccc}
   &\includegraphics[width=0.1\textwidth]{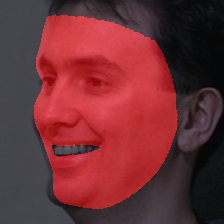}
   &\includegraphics[width=0.1\textwidth]{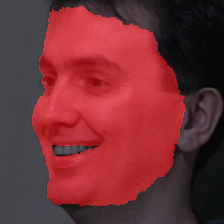}
   &\includegraphics[width=0.1\textwidth]{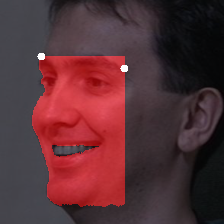}\\
   &\footnotesize{(a) Initial mask} 
   &\footnotesize{(b) After filtering}
   &\footnotesize{(c) After cropping}
\end{tabular}
   \caption{The mask processing for an observed image. The initial mask is essentially the texture sampling regions. It is then filtered using a joint edge-preserving filering with the image as guidance. The final mask (c) is obtained by excluding occluded regions using 2D detected landmarks on eyebrows (the white points).}
   \vspace{-5pt}
\label{fig:dtfilter}
\end{figure}

Till now, we are discussing the rendered projections without considering occlusions. 
To exclude occluded facial regions, we employ visibility masks to obtain  $\mathcal{M}$. Note that Eq. \eqref{eq.consistphoto} is for the ideal case, where the visibility mask is the same for both rendered image and observed image. In practice, with imperfect 3DMM and pose parameters, we need different masks for rendered image and observed image to enforce the photometric consistency (see Sec. \ref{sec:selfsup} for details). For rendered image, we simply extract a visibility mask by excluding regions that may be occluded in other views using 3D vertices corresponding to 2D facial landmarks (the correspondences between 3D vertices and 68-points 2D facial landmarks are provided by \cite{zhu2016face}). Fig. \ref{fig:masks} illustrates an example of the visibility masks for all three views. 
For the observed real image, we obtain an initial mask using the texture sampling regions. Then a joint edge-preserving filtering \cite{gastal2011domain} is performed on the initial mask, with the input real image as guidance, to force the edges of the mask aligned well with the facial regions of the input image. Finally the regions that may be occluded in other views are excluded using 2D detected landmarks, similar to the processing of masks for rendered images  (see Fig. \ref{fig:dtfilter}). Note that for the frontal observed image, there are two different visibility masks when viewed from left and right sides, respectively. We denote the set of pixels in the corresponding masks as $\mathcal{M}_B^{(A)}$ and $\mathcal{M}_B^{(C)}$. 

\subsection{Losses and Training}\label{sec:losses}

In order to obtain a good initialization and avoid trapping into local minima, we first pretrain the CNN using supervised labels on the 300W-LP dataset \cite{zhu2016face}, where ground-truth 3DMM and pose parameters are obtained via conventional 3DMM fitting algorithms and multi-view images are generated by face profiling augmentation. After the pretraining  converges, we then perform self-supervised training on the Multi-PIE dataset \cite{gross2010multi}, where multi-view facial images are taken in controlled indoor settings.
The training losses are detailed in the following section.

\subsubsection{Supervised Pretraining}\label{sec:supervised}

In supervised pretraining, the ground-truth landmarks, 3DMM and pose parameters are provided. In the dataset 300W-LP, for each real facial image, several synthetic rendered views are generated. During the training stage, we randomly select a set of multi-view images for each face, which contains left, frontal, and right views. We use ground-truth landmarks, 3DMM and pose parameters as supervision, as well as regularizations on 3DMM parameters. The supervised training loss is
\begin{equation}\label{eq.superviseloss}
  L_{\text{sup}} = \lambda_1 L_{\text{landmark}} + \lambda_2 L_{\text{pose}} + \lambda_3 L_{\text{3DMM}} + \lambda_4 L_{\text{reg}},
\end{equation}
where $L_{\text{landmark}}$ is the landmark alignment loss similar to \cite{tewari2017mofa}, $L_{\text{pose}}$ and $L_{\text{3DMM}}$ are L2 losses between predictions and  ground-truths, $L_{\text{reg}}$ is the regularization loss on 3DMM parameters also similar to \cite{tewari2017mofa}. The weighting $\lambda_{1,2,3,4}$ are hyper-parameters controlling the trade-off between losses.

\subsubsection{Self-supervised Training}\label{sec:selfsup}

During the self-supervised training stage, we enforce the photometric consistency between observed image and synthetic rendered image to incorporate multi-view geometric constraints. 
From Eq. \eqref{eq.consistphoto} we derive the photometric loss 
\begin{equation}\label{eq.photoloss}
  L_{\text{photo}}(\mathbf{I}_B, \mathbf{I}_{A \rightarrow B}) = 
  \sum\limits_{\mathbf{u} \in \mathcal{M}_B^{(A)} \cup \mathcal{M}_{A \rightarrow B}} \| \mathbf{I}_{B}[\mathbf{u}] - \mathbf{I}_{A \rightarrow B}[\mathbf{u}] \|^2_2,
\end{equation}
where $\mathcal{M}_B^{(A)}$ and $\mathcal{M}_{A \rightarrow B}$ are the sets of pixels in visibility masks for $\mathbf{I}_{B}$ (viewed from the left side) and $\mathbf{I}_{A \rightarrow B}$ respectively. 
Note that here we use the union of $\mathcal{M}_B^{(A)}$ and $\mathcal{M}_{A \rightarrow B}$ such that misalignment errors can be taken into considerations. Unfortunately, we find that using only the photometric loss could lead to bad alignment in practice. The reason is that the pixels within facial regions are similar to each other such that mis-matching easily happens. In order to increase the reliability of the dense correspondences between observed image and rendered image, we introduce an additional novel  alignment loss into the training.

\begin{figure}
    \centering
    \includegraphics[width=0.5\textwidth]{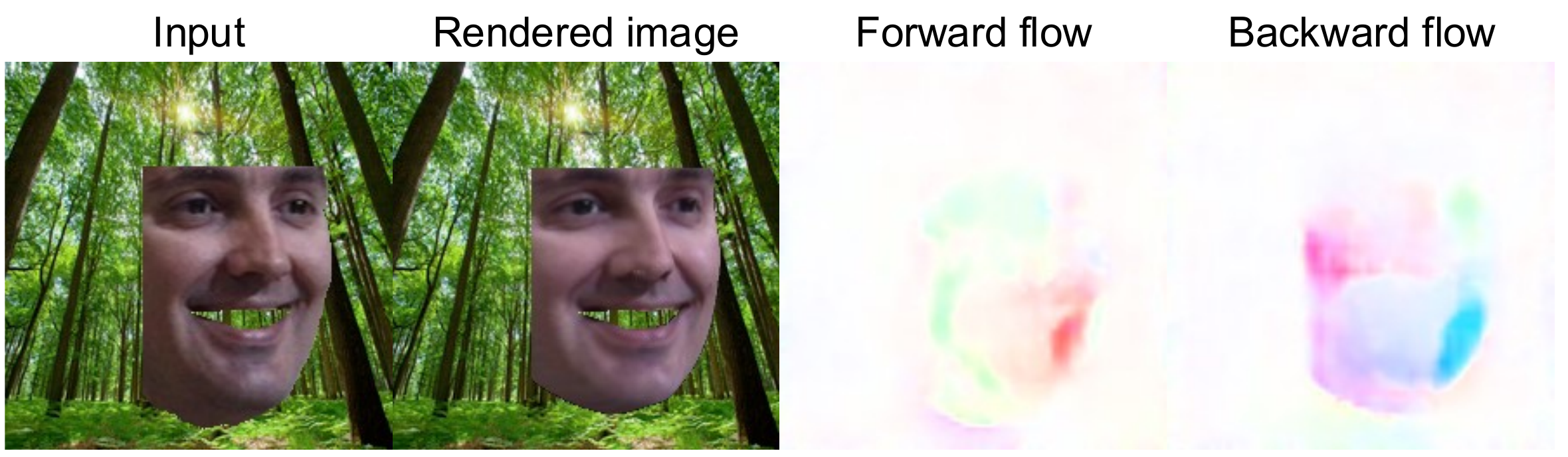}
    \caption{Optical flows between observed and rendered images.}
    \label{fig:flow}
    \vspace{-10pt}
\end{figure}

We employ a differentiable dense optical flow estimator
to compute the flow between observed image and rendered image, and then use the sum of squared flow magnitudes at all pixels as the alignment loss. Since the dense optical flow estimator tends to estimate smoothed flow fields, individual mis-matchings can be largely suppressed. For example, to enforce the photometric consistency between $\mathbf{I}_B$ and $\mathbf{I}_{A \rightarrow B}$, we compute the alignment loss as
\begin{equation}\label{eq.viewalignloss}
  L_{\text{align}}(\mathbf{I}_B, \mathbf{I}_{A \rightarrow B}) = \arrowvert \mathbf{F}(\mathbf{I}_B, \mathbf{I}_{A \rightarrow B}) \arrowvert + \arrowvert \mathbf{F}(\mathbf{I}_{A \rightarrow B}, \mathbf{I}_B)\arrowvert,
\end{equation}
where $\mathbf{F}(\cdot)$ denotes the optical flow estimator. Note that here bi-directional optical flows are employed. Besides, in order to reduce the distractions of optical flow estimation errors in uninterested regions, we fill in the the regions outside visibility masks with textures whose flow can be easily estimated (see Fig. \ref{fig:flow} for an example).

For the three-view setting, we compute the photometric loss and alignment loss between 4 pairs of images: $(\mathbf{I}_B, \mathbf{I}_{A \rightarrow B})$, $(\mathbf{I}_B, \mathbf{I}_{C \rightarrow B})$, $(\mathbf{I}_A, \mathbf{I}_{B \rightarrow A})$, and $(\mathbf{I}_C, \mathbf{I}_{B \rightarrow C})$. Additionally, to increase the training stability, we also adopt the landmark loss $L_{\text{landmark}}$ during self-supervised training, where the landmarks are detected via a state-of-the-art landmark detector from \cite{bulat2017far} automatically. To sum, the self-supervised training loss is 
\begin{equation}\label{eq.selfsuperviseloss}
  L_{\text{self-sup}} = \lambda_5 L_{\text{landmark}} + \lambda_6 L_{\text{photo}} + \lambda_7 L_{\text{align}},
\end{equation}
where both photometric loss $L_{\text{photo}}$ and alignment loss $L_{\text{align}}$ are computed from the above 4 pairs of images. The hyper-parameters $\lambda_{5,6,7}$ control the trade-off between losses.

\section{Experiments}
In this section, we first introduce the datasets, evaluation metrics, and implementation details for conducting the experiments (Sec. \ref{sec:datasets} and \ref{sec:implementdetail}). 
We then demonstrate the effectiveness of the proposed approach with extensive ablation studies in Sec. \ref{sec:ablation}. 
Finally, quantitative and qualitative comparisons to state-of-the-art single-view 3DMM-based approaches are presented in Sec. \ref{sec:comparison}.

\subsection{Datasets and Metrics}\label{sec:datasets}
\textbf{Training Datasets.} 1) Our supervised pretraining is performed on  300W-LP dataset \cite{zhu2016face}, which contains over 60,000 images derived from  3,837 face images by varying poses using face profiling synthesis method \cite{zhu2016face}. Ground-truth landmarks, 3DMM and pose parameters are provided by the dataset.
We sample triplet consists of a front, left, and right view image from 300W-LP dataset using the provided yaw angles, which results in 140k training triplets in total. 

2) Our self-supervised training is performed on Multi-PIE dataset \cite{gross2010multi}, which contains over 750,000 images recorded from 337 subjects using 15 cameras in different directions under various lighting conditions. 
We take frontal-view images as anchors and randomly select side-view images (left or right) to get 50k training triplets and 5k testing triplets, where the subjects in testing split do not appear in training split. 
Note that whether an image is in frontal, left, or right view can be determined by the provided camera ID.

\textbf{Evaluation Datasets.} 1) We mainly perform quantitative and qualitative evaluations on the MICC Florence dataset  \cite{Bagdanov:2011:FHF:2072572.2072597}, which consists of 53 identities of persons with neutral expression and ground-truth 3D scans are available. Each person contains three videos of ``indoor-cooperative", ``indoor", and ``outdoor" respectively. 
To experiment with the multi-view setting addressed in this paper, we mannually select a set of multi-view frames for each person, such that his/her expressions are consistent in different views. Since it is difficult to select such sets of frames in the ``outdoor'' videos, we only perform evaluations on the ``indoor-cooperative" and ``indoor" videos.
2) Qualitative evaluations are further performed on Color FERET dataset \cite{phillips2000feret,phillips1998feret} and MIT-CBCL face recognition database \cite{weyrauch2004component}, where multi-view facial images are available.

\textbf{Evaluation Metrics.} In the quantitative evaluations on MICC dataset, we follow the evaluation metrics from \cite{genova2018unsupervised}, which compute point-to-plane L2 errors between predict 3D models and ground-truth 3D scans. Here, we abandon subjects of ID 2 and 27 as their ground-truth 3D scans are flawed and also excluded in other work \cite{tewari2017mofa,genova2018unsupervised}. 

\subsection{Implementation Details}\label{sec:implementdetail}

We use PWCNet \cite{sun2018pwc} as our differentiable optical flow estimator in the self-supervised training step. Note that during our training, the weights of PWCNet is fixed. 
We crop input images according to bounding boxes of facial landmarks (either ground-truth or detected with \cite{bulat2017far}) and resize them to 224$\times$224. 
To augment the training data, we add random shift with 0$\sim$0.05 of input size to the bounding box. 
We adopt Adam \cite{kingma2014adam} as the optimizer.
The batchsize is set to 12.
The supervised pretraining is trained on 300W-LP for 10 epoches with learning rate 1e-5, and the self-supervised training is trained on Multi-PIE for 10 epoches with learning rate 1e-6. 
The default weights for balancing losses are set to $\lambda_1=0.1$, $\lambda_2=10$, $\lambda_3=1$, $\lambda_4=1$, $\lambda_5=1$, $\lambda_6=10$, and $\lambda_7=0.1$. We set different weights for different loss terms to make their numbers in a similar scale. The weights $\lambda_1$ and $\lambda_7$ are set to relatively smaller values as they represent pixel distances. The weights $\lambda_2$ and $\lambda_6$ are set to larger values as pose parameters and pixel values of input images are normalized to $[0,1]$.

\begin{table}[!t]\small

  \setlength{\tabcolsep}{3.0pt}
  \begin{center}
      \begin{tabular}{lccccccc}
      \toprule
      \multirow{2}{*}{Ours}&
      \multicolumn{3}{c}{Self-supervised Loss}&
      \multicolumn{2}{c}{INC}&
      \multicolumn{2}{c}{IND}\\ 
      &$L_{\text{landmark}}$ & $L_{\text{photo}}$& $L_{\text{align}}$ & Mean & Std & Mean & Std\\
      \midrule
      v1 & -- & -- & -- 
      &1.266 & 0.297 & 1.252 & 0.285\\
      v2 &$\surd$ & $\surd$& $\times$
      &1.240 & 0.258 & 1.252 & 0.245\\
      v3 &$\surd$ & $\times$& $\surd$
      &1.227 & 0.248 & 1.245 & 0.240\\
      v4 &$\surd$ & $\surd$& $\surd$
      &\textbf{1.220} & \textbf{0.247} & \textbf{1.228} & \textbf{0.236}\\
      \bottomrule
      \end{tabular}
  \end{center}
      \caption{Mean error of our approach on the MICC dataset. The versions: v1 for the supervised pretrained model; v2-v4 for the self-supervised trained model with different losses.}
      \label{tab:ablationstudy}
\end{table}

\begin{figure}[!t]
    \vspace{-5pt}
    \centering
    \includegraphics[width=\linewidth]{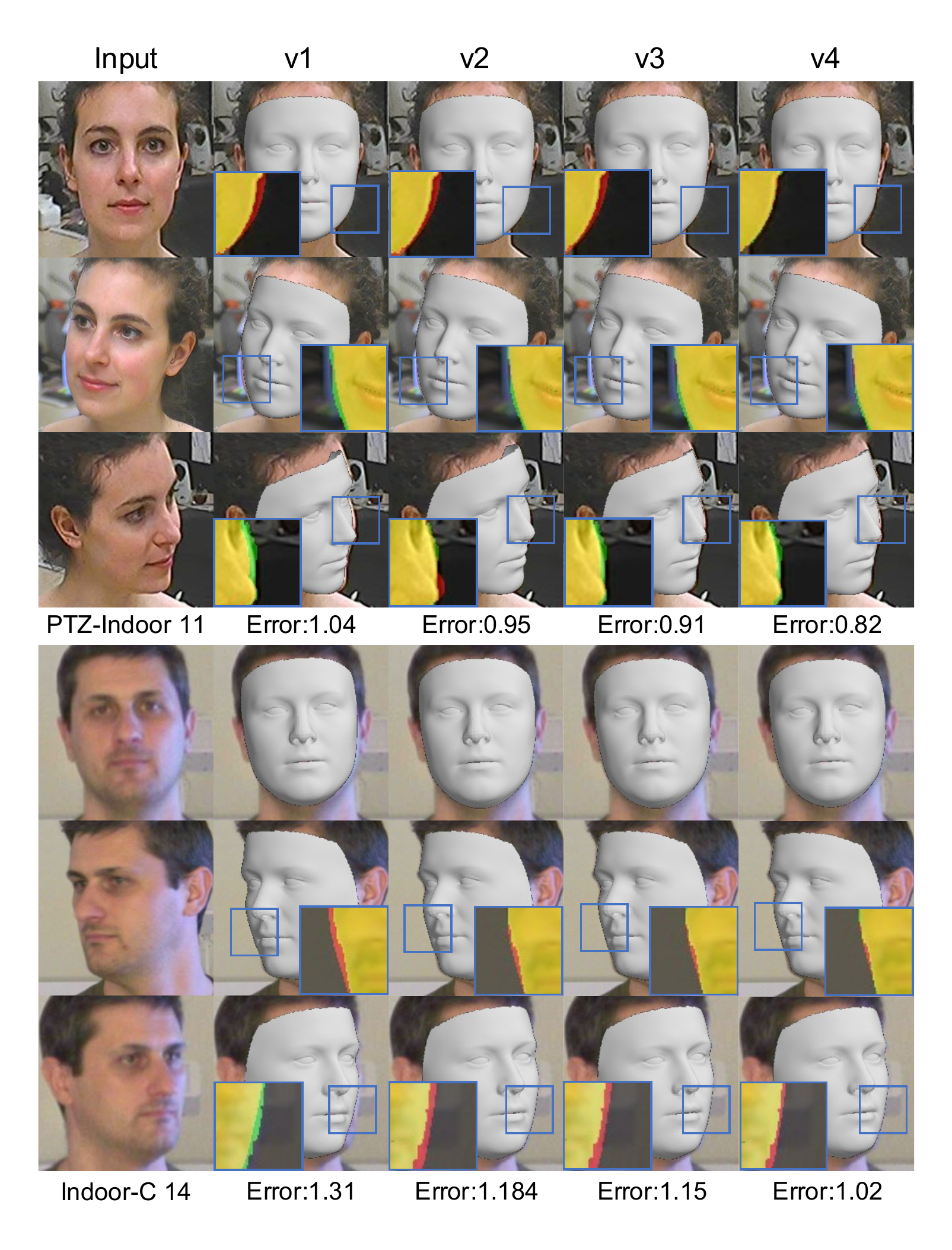}
    \caption{Visual examples of ablation study on the MICC dataset. The meanings of the colors in the close-ups are as follows. Red: the projection area from 3D to 2D exceeds the observed facial boundary. Green: the projection area is smaller than the facial area. Yellow: overlap between projection and facial areas.}
    \label{fig:ablation}
    \vspace{-5pt}
\end{figure}

\begin{figure*}[!t]
    \vspace{-10pt}
    \centering
    \includegraphics[width=\textwidth]{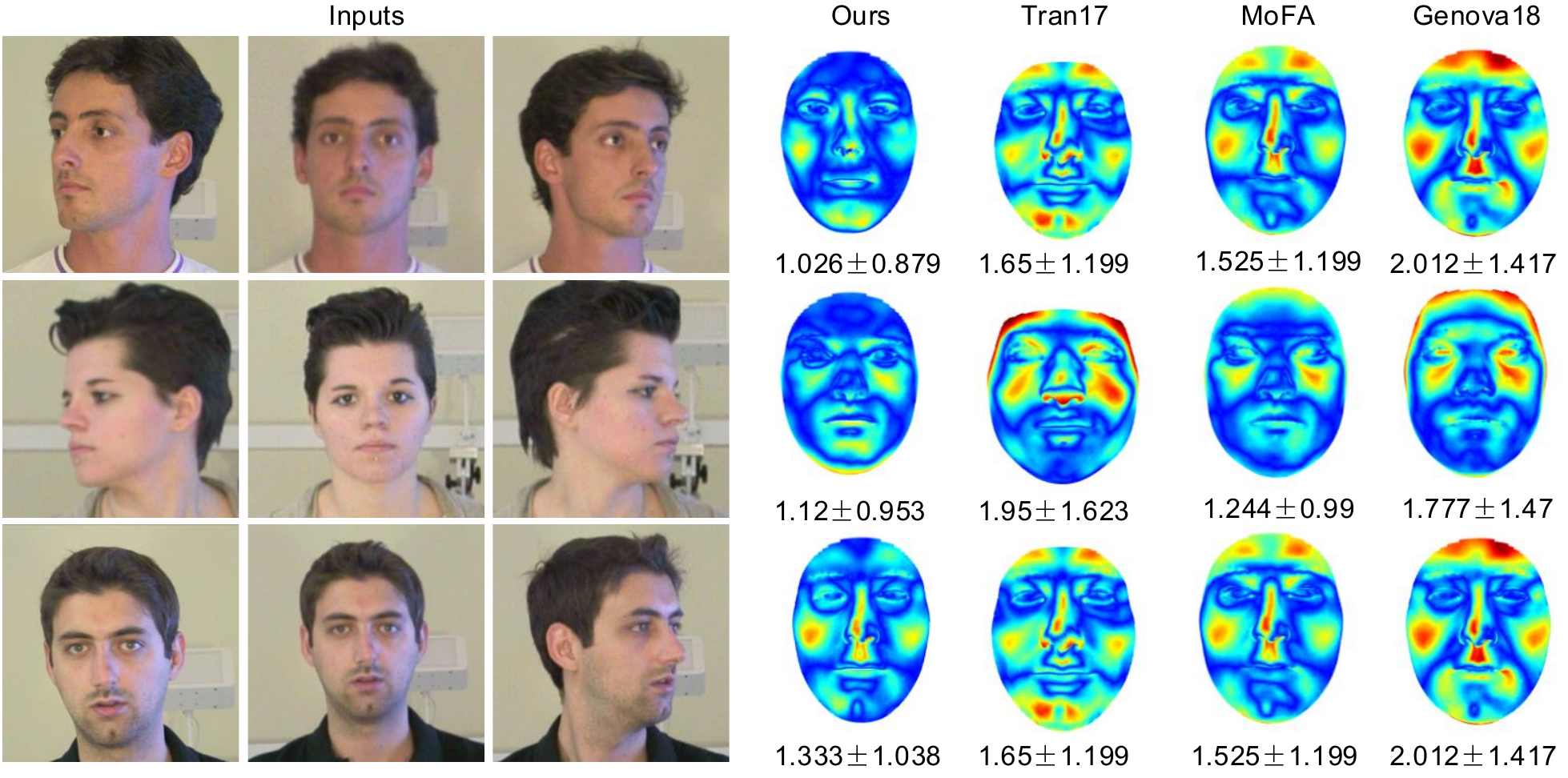}
    \caption{Examples of error map comparison on the MICC dataset.}
    \label{fig:heatmapcomp}
    \vspace{-5pt}
\end{figure*}

\subsection{Ablation Study}\label{sec:ablation}

We conduct a series of experiments on MICC dataset to demonstrate the effectiveness of each component in our approach. 
Table \ref{tab:ablationstudy} shows the mean errors of different versions of our model. 
From the results we observe that, compared with the supervised pretrained model (v1), the self-supervised trained model with only photometric loss (v2) reduces the mean error by 0.026 for ``indoor-cooperative'' but none for ``indoor'' images, while the model with only alignment loss (v3) reduces the mean error by 0.039 for ``indoor-cooperative'' and 0.007 for ``indoor'' images, which is a moderate improvement over photometric loss. Combining the photometric loss and alignment loss (v4) gives the best results, an error reduction of 0.046 and 0.024.

Fig. \ref{fig:ablation} shows two visual examples of the ablation study. From the close-ups we can clearly observe the performance improvements from v1 to v4. Specifically, take the right-side view of the bottom person as an example, we can observe that the facial silhouette of the input face is flat, while in the result from v1 it seems a little bit plump and it becomes much more flatter in the result from v4. 
The same trends can be found in other examples by inspecting the alignment of 3D models to the facial silhouettes.

\begin{figure}
    \vspace{-5pt}
    \centering
    \includegraphics[width=\linewidth]{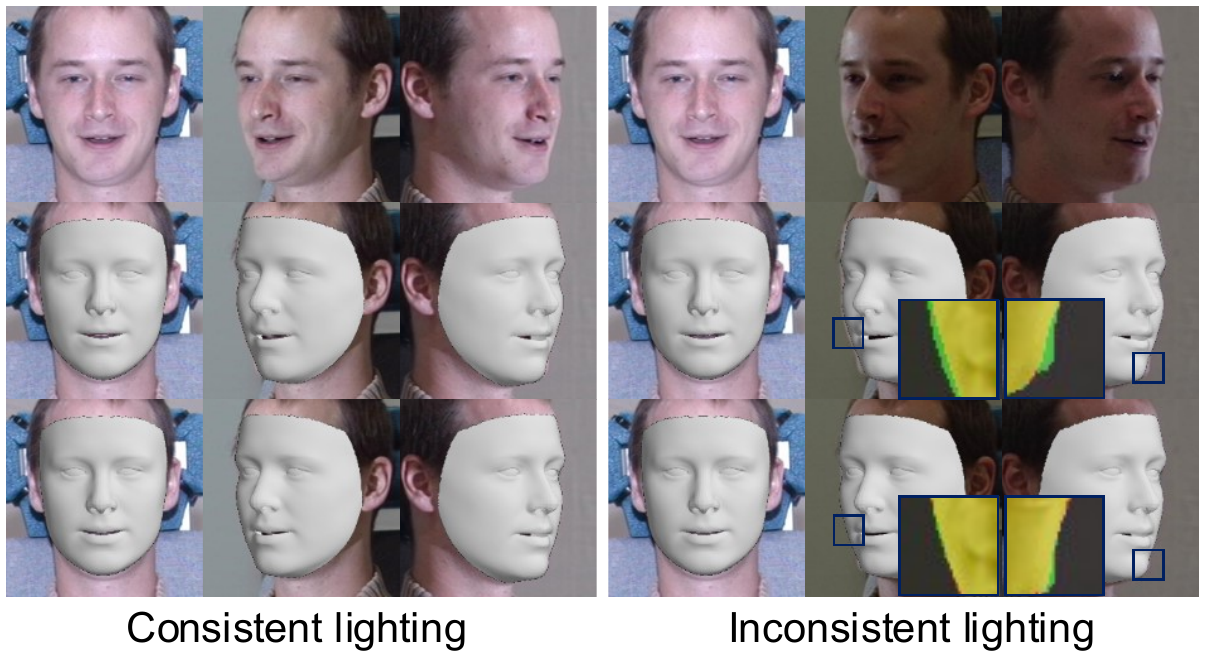}
    \caption{Experiments on inconsistent lighting conditions across views. First row: input. Second row: results obtained with only the photometric loss. Third row: results obtained with both photometric loss and alignment loss.}
    \label{fig:invariant}
    \vspace{-15pt}
\end{figure}

We further conduct studies under varying lighting conditions across views to demonstrate the effectiveness of the proposed alignment loss to handle lighting changes. Fig. \ref{fig:invariant} shows an example. In this example, when the lighting is consistent across the three views (left), the model trained with only photometric loss performs almost as good as the model trained with both photometric loss and alignment loss. But when the lighting is inconsistent across the views, the result obtained from only photometric loss is much worse than that from both losses. 
The reason why the alignment loss is robust to lighting changes is due to the optical flow estimator, which is already trained to deal with lighting changes of input images. 

\subsection{Comparisons to State-of-the-art Methods}\label{sec:comparison}

\begin{table}[!t]\small

  \setlength{\tabcolsep}{3pt}
  \begin{center}
      \begin{tabular}{ccccc}
      \toprule
      \multirow{2}{*}{Method}&
      \multicolumn{2}{c}{INC}&
      \multicolumn{2}{c}{IND}\\ 
      & Mean & Std & Mean & Std\\
      \midrule
      Tran \etal \cite{tran2017regressing}
      &1.443&0.292&1.471&0.290\\
      Tran \etal + pool
      &1.397&0.290&1.381&0.322\\
      Tran \etal + \cite{piotraschke2016automated}
      &1.382 & 0.272 & 1.430 & 0.306\\
      MoFA \cite{tewari2017mofa}
      &1.405 & 0.306 &1.306 & 0.261 \\
      MoFA + pool
      &1.370 & 0.321 &1.286 & 0.266 \\
      MoFA + \cite{piotraschke2016automated}
      &1.363  & 0.326 & 1.293 & 0.276\\
      Genova \etal \cite{genova2018unsupervised}
      & 1.405 & 0.339 & 1.271 & 0.293\\
      Genova \etal + pool
      & 1.372 & 0.353 & 1.260 & 0.310\\
      Genova \etal + \cite{piotraschke2016automated}
      & 1.360 & 0.346 & 1.246 & 0.302\\
      Ours
      &\textbf{1.220} & \textbf{0.247} & \textbf{1.228} & \textbf{0.236}\\
      \bottomrule
      \end{tabular}
  \end{center}
      \caption{Comparison of mean error on the MICC dataset.}
      \label{tab:quantcomp}
\end{table}

\begin{figure*}
    \vspace{-10pt}
    \centering
    \includegraphics[width=\textwidth]{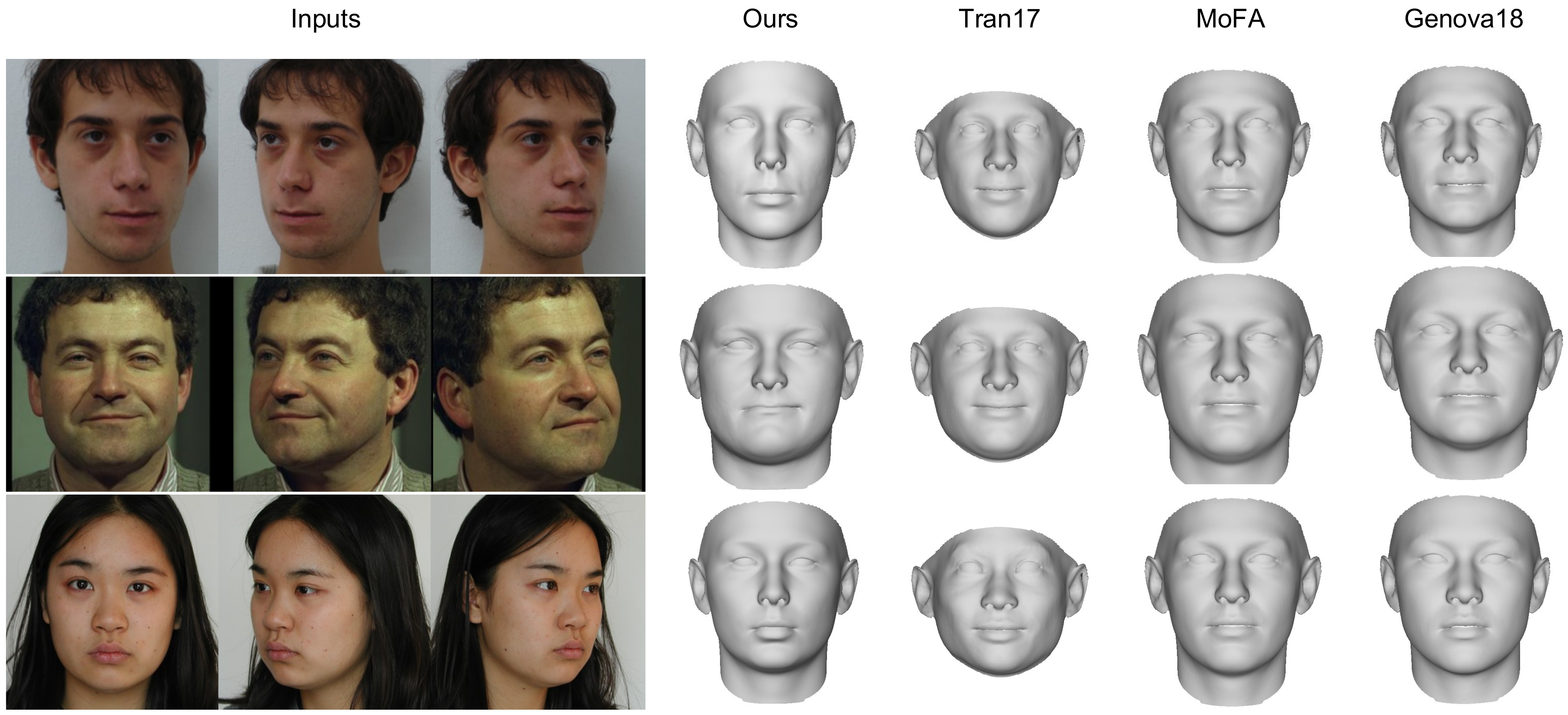}
    \caption{Examples of visual comparison with the other methods. More examples are in supplementary materials.}
    \label{fig:visualcomparison}
    \vspace{-10pt}
\end{figure*}

\begin{figure}
    \centering
    \includegraphics[width=\linewidth]{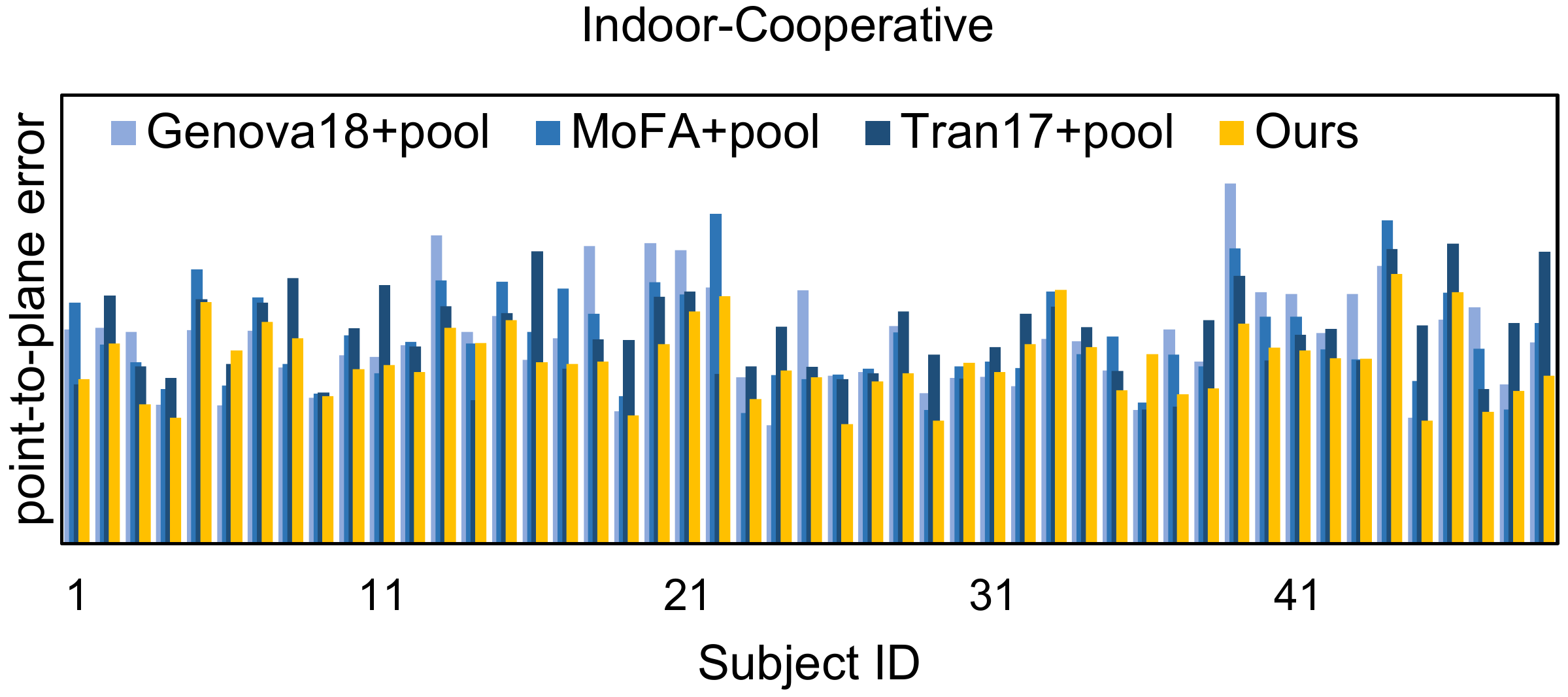}
    \includegraphics[width=\linewidth]{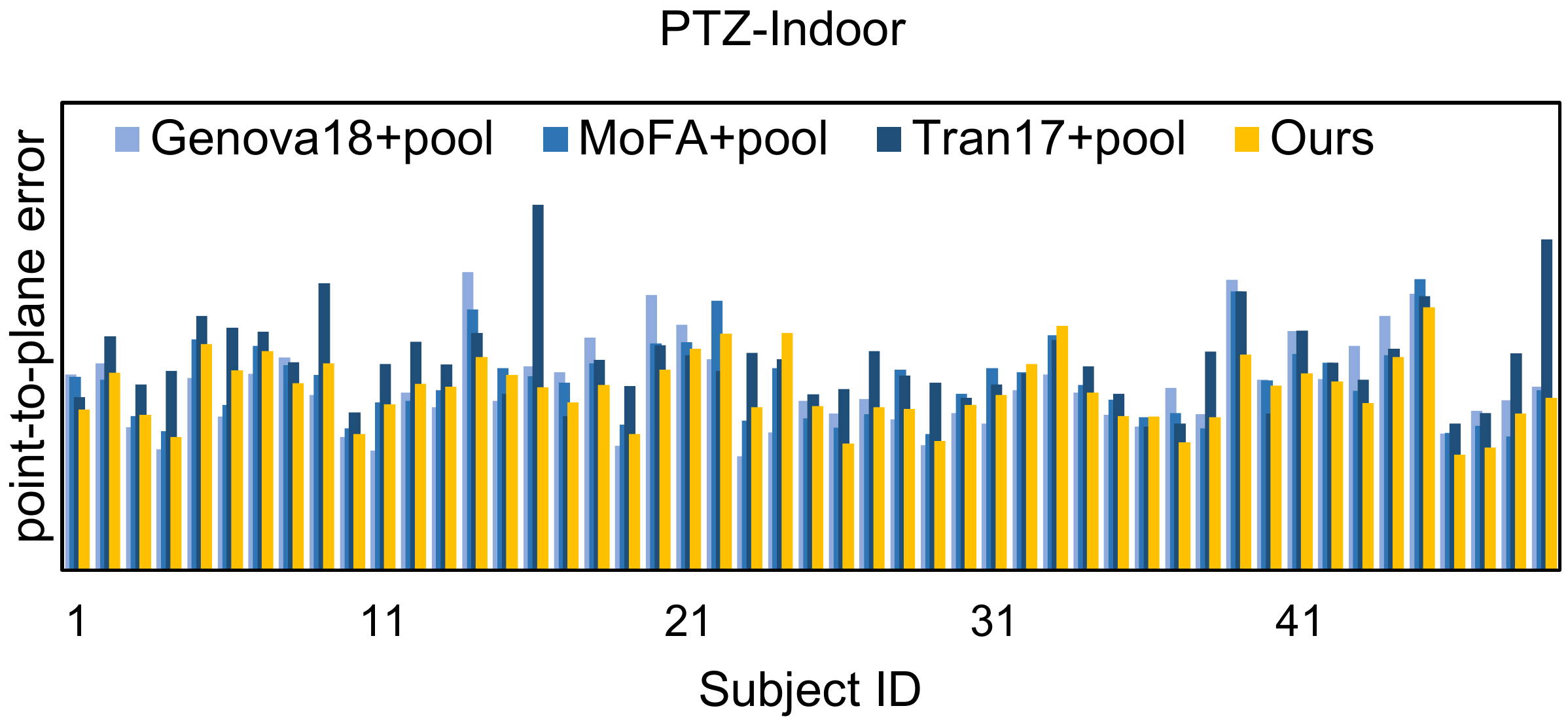}
    \caption{Detailed comparisons for each subject in MICC dataset.}
    \label{fig:micc_each_id}
    \vspace{-10pt}
\end{figure}

We first compare our results on MICC dataset with state-of-the-art single-view 3DMM reconstruction methods. 
To evaluate single-view methods on our three-view evaluation triplets for each person, we first use their model to predict 3D model a 3D model for each input image. Then three different evaluation settings are employed to ensure fair comparisons. The first one is to calculate the point-to-plane errors for each 3D model and then average the errors. The second one is to average the three predicted 3D models in a triplet and then compute the point-to-plane errors between the pooled 3D model with ground-truth model (shown in Table \ref{tab:quantcomp} as ``+pool'' entries). The third one is to compute the weighted average of three predicted 3D models as \cite{piotraschke2016automated} and then compute the point-to-plane errors (shown in Table \ref{tab:quantcomp} as ``+\cite{piotraschke2016automated}'' entries). Table \ref{tab:quantcomp} shows the mean errors of the comparison. 
The proposed method outperforms all single-view methods in both settings. 
Fig. \ref{fig:micc_each_id} shows the detailed numerical comparisons for each subject in the dataset. 
Several examples of the comparison of detailed error maps are presented in Fig. \ref{fig:heatmapcomp}.

\begin{figure}
    \centering
    \includegraphics[width=\linewidth]{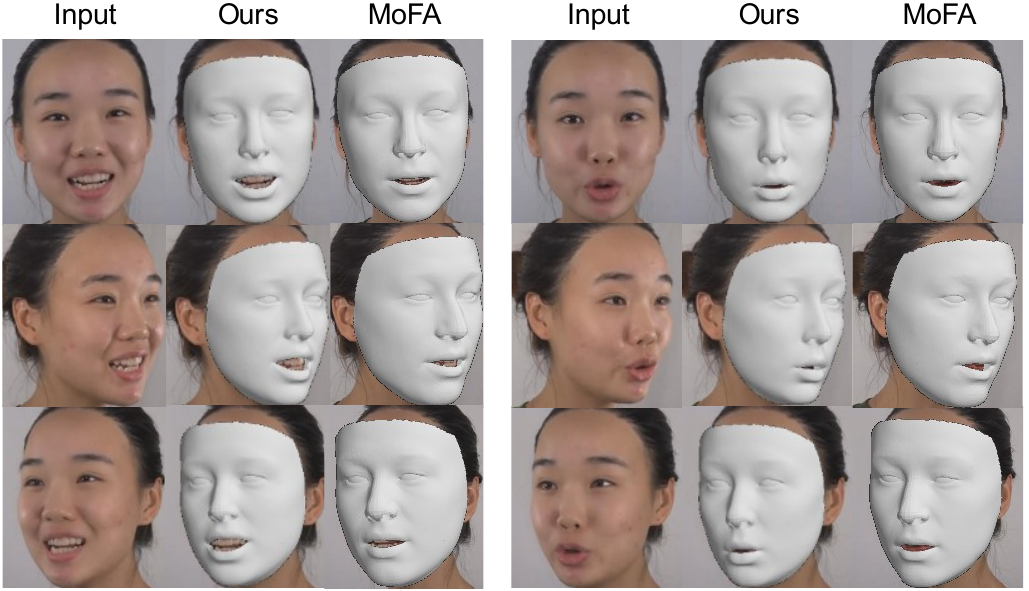}
    \caption{Examples of visual comparison to MoFA in different facial expressions. Our method can produce more accurate shapes and expressions. More examples are in supplementary materials.}
    \label{fig:mofacomp}
    \vspace{-10pt}
\end{figure}

We further present some visual comparisons using images from other datasets such as Color FERET dataset \cite{phillips2000feret,phillips1998feret} and MIT-CBCL face recognition database \cite{weyrauch2004component}, where multi-view facial images are available. 
Fig. \ref{fig:visualcomparison} shows several examples of the visual comparisons to single-view methods in neutral expression. 
Fig. \ref{fig:mofacomp} shows several examples of the visual comparisons to MoFA in different facial expressions. 
The superiority of our method over single-view methods can be observed in these comparisons.

\section{Conclusions}

In this paper, we presented a novel approach to regress 3DMM parameters from multi-view facial images with an end-to-end trainable CNN. 
Different from single-view 3DMM-based CNNs, our approach explicitly incorporates multi-view geometric constraints as the photometric loss and alignment loss between different views with the help of rendered projections via predicted 3D models. 
The alignment loss was computed via a differentiable dense optical flow estimator, which enables the flow errors to backpropagate to the 3DMM parameters to be predicted. 
The effectiveness of the proposed approach was validated through the extensive experiments. 
Our study essentially explores model-based multi-view reconstruction using deep learning, which we believe will inspire more future research.

{\small
\bibliographystyle{ieee_fullname}
\bibliography{egbib}
}

\end{document}